\documentclass[a4paper]{article}
\usepackage{hyperref}
\usepackage{INTERSPEECH2021}

\title{Alzheimer's Dementia Recognition Using Acoustic, Lexical, Disfluency and Speech Pause Features Robust to Noisy Inputs}
\name{Morteza Rohanian$^1$, Julian Hough$^1$, Matthew Purver$^{1,2}$}
\address{
$^1$Cognitive Science Group\\School of Electronic Engineering and Computer Science\\
  Queen Mary University of London\\
  $^2$Department of Knowledge Technologies, Jožef Stefan Institute}
\email{ \{m.rohanian, j.hough, m.purver\}@qmul.ac.uk}

\begin{document}

\maketitle
\begin{abstract}
  We present two multimodal fusion-based deep learning models that consume ASR transcribed speech and acoustic data simultaneously to classify whether a speaker in a structured diagnostic task has Alzheimer's Disease and to what degree, evaluating the ADReSSo challenge 2021 data. Our best model, a BiLSTM with highway layers using words, word probabilities, disfluency features, pause information, and a variety of acoustic features, achieves an accuracy of 84\% and RSME error prediction of 4.26 on MMSE cognitive scores. While predicting cognitive decline is more challenging, our models show improvement using the multimodal approach and word probabilities, disfluency, and pause information over word-only models. We show considerable gains for AD classification using multimodal fusion and gating, which can effectively deal with noisy inputs from acoustic features and ASR hypotheses.

\end{abstract}


\section{Introduction}

Alzheimer's disease (AD) is a chronic neurodegenerative disease that affects memory, language, cognitive skills, and the ability to perform simple everyday tasks. 

Throughout the course of AD, patients have been observed suffering a loss of lexical-semantic skills, including suffering anomia, reduced word comprehension, object naming problems, semantic paraphasia, and a reduction in vocabulary and verbal fluency \cite{forbes2005detecting,bayles1982potential}. Speech in patients with AD is mostly characterised by a low speech rate and frequent hesitations at the phonetic and phonological level; however, the syntactic ability is better preserved than lexical-semantic ability in AD patients at the early stages of the disease\cite{kave2003morphology}.

The presence of cognitive dysfunction must be confirmed by neuropsychological tests such as the mini-mental state assessment (MMSE) performed in medical clinics before an AD diagnosis can be made. The existence of typical neurological and neuropsychological characteristics and a clinical examination of the patient's history are used to make a diagnosis.

Detecting early diagnostic biomarkers that are non-invasive and cost-effective is of great value for clinical assessments. Several previous studies have investigated AD diagnosis via acoustic, lexical, syntactic, and semantic aspects of speech and language. More interactional aspects of language, like disfluencies, and purely non-verbal features, such as intra- and inter-speaker silence, can be key features of AD conversations. If useful for diagnosis, these features can have many advantages: they are easy to extract and are relatively language, subject, and task agnostic.

In terms of speech features, the number of pauses, pause proportion, phonation time, phonation–to–time ratio, speech rate, articulation rate, and noise–to–harmonic ratio were all found to be related to the severity of Alzheimer's disease \cite{szatloczki2015speaking}. Weiner et al. \cite{weiner2016speech} used a Linear Discriminant Analysis (LDA) classifier with a set of acoustic features including the mean of silent segments, silence durations, and silence-to-speech ratio to differentiate subjects with AD from the control group, achieving an 85.7\% AD binary classification. Ambrosini et al. \cite{ambrosini2019automatic} used selected acoustic features (pitch, voice breaks, shimmer, speech rate, syllable duration) to detect mild cognitive impairment from a spontaneous speech task.

Lexical features from spontaneous speech have been shown to be informative in terms of features that assist AD detection. For example, Jarrold et al. \cite{jarrold2014aided} merged acoustic features with the frequency occurrence of 14 distinct parts of speech features. Abel et al. \cite{abel2009connectionist} modeled patient speech errors (naming and repetition disorders) to aid AD diagnosis.

Modeling multimodal input for AD detection has also been studied. Gosztolya et al. \cite{gosztolya2019identifying} looked at how two SVM models with different sets of acoustic and linguistic features could be combined. Their research demonstrated how audio and lexical features could provide additional knowledge about an individual with AD.

Among other similar tasks within cognitive state prediction like depression, research has been done on integrating temporal information from two or more modalities using multimodal fusion \cite{rohanian2019detecting}. The different predictive capacities of each modality and their different levels of noise are a major challenge for these models. A gating mechanism is effective in controlling the level of contribution of each modality to the final prediction in a variety of multimodal tasks, including in AD classification and regression \cite{Rohanian2020}.

This paper constitutes an entry into the Alzheimer's Dementia Recognition through Spontaneous Speech (ADReSSo) challenge 2021 \cite{bib:LuzEtAl21ADReSSo}, which involves an AD classification and MMSE score regression tasks, in addition to a cognitive decline (disease progression) inference task using only the audio data from formal diagnosis interviews with patients as input. In the first two tasks, participants are required to rate the severity of Alzheimer's disease in various subjects, with the target severity determined by their MMSE scores. In the third task, participants should identify those patients who exhibit cognitive decline within two years.

In this paper, we were particularly interested in the benefit of fusing ASR results (rather than transcripts) with acoustic data and whether self-repair disfluencies and unfilled pauses in individuals' speech and language model probabilities (a measure of lexical predictability) from automatic speech recognition (ASR) results would help predict the severity of the patient's cognitive impairment.

Inspired by \cite{Rohanian2020}, to detect AD, we used audio and text features to model the sessions in a Bidirectional Long-Short Term Memory (BiLSTM) neural network. We used the Bidirectional Encoder Representations from Transformers (BERT) model to classify AD from speech recognition results in a separate experiment. Our findings suggest that AD can be identified using pure sequential modelling of the speech recognition results from the interview sessions with limited details of the structure of the description tasks. Disfluency markers, unfilled pauses, and language model probabilities were also found to have predictive power for detecting Alzheimer's disease.

\section{Data and features}

Two distinct datasets were used for the ADReSSo Challenge:

\begin{enumerate}

    \item a set of speech recordings of picture descriptions produced by both patients with an AD diagnosis and subjects without AD (controls), who were asked to describe the Boston Diagnostic Aphasia Exam's Cookie Theft picture \cite{bib:LuzEtAl21ADReSSo}.
        \item  a set of speech recordings of Alzheimer’s
patients performing a category (semantic) fluency task
\cite{benton1968differential} at their baseline visit for prediction of cognitive decline over two years.

\end{enumerate}

Dataset 1 for AD classification and severity detection includes 237 audio recordings, and the state of the subjects is assessed based on the MMSE score. MMSE is a commonly used cognitive function test for older people. It involves orientation, memory, language, and visual-spatial skills tests. Scores of 25-30 out of 30 are considered as normal, 21-24 as mild, 10-20 as moderate, and $<$10 as a severe impairment. 

Dataset 2 for the disease prognostics task (prediction of
cognitive decline) was created from a longitudinal cohort study
involving AD patients. The period for assessing disease progression spanned the baseline and the year-2 data collection visits of the patients to the clinic. The task involves
classifying patients into `decline' or ’no-decline’ categories,
given speech collected at baseline as part of a verbal fluency
test.

Various features were extracted automatically from both datasets for the 3 ADReSSo tasks as described below.

\subsection{Acoustic features}

A set of 79 audio features were extracted using the COVAREP acoustic analysis framework software, a package used for automatic extraction of features from speech \cite{degottex2014covarep}. We sampled the audio features at 100Hz and used the higher-order statistics (mean, maximum, minimum, median, standard deviation, skew, and kurtosis) of COVAREP features. The features include \textit{prosodic features} (fundamental frequency and voicing), \textit{voice quality features} (normalized amplitude quotient, quasi-open quotient, the difference in amplitude of the first two harmonics of the differentiated glottal source spectrum, maxima dispersion quotient, parabolic spectral parameter, spectral tilt/slope of wavelet responses, and shape parameter of the Liljencrants-Fant model of the glottal pulse dynamics) and \textit{spectral features} (Mel cepstral coefficients 0-24, Harmonic Model and Phase Distortion mean 0-24 and deviations 0-12). Segments without audio data were set to zero. A standard zero-mean and variance normalization was applied to features. We omitted all features with no statistically significant univariate correlation with the results of the training set.

\subsection{Linguistic Features}

For automatically transcribing the audio files, we used the free trial version of IBM's Watson Speech-To-Text service.\footnote{\url{https://www.ibm.com/uk-en/cloud/watson-speech-to-text}} The service offers ASR on the audio data which has considerable noise and may be affected by non-standard North American dialect of the patients - the average Word Error Rate (WER) on 10 transcripts we randomly selected from the training data is 32.8\%. The Watson service, crucially for our task, does not filter out hesitation markers or disfluencies \cite{baumann2017recognising}. It also outputs word timings that we use as features in our system.

For our models which did not use BERT, a pre-trained GloVe model \cite{pennington2014glove} was used to extract the lexical feature representations from the picture description transcript and convert the utterance sequences into word vectors. We selected the hyperparameter values, which optimised the output of the model on the training set. The optimal dimension of the embedding was found to be 100.

\subsection{Disfluencies}

 Disfluencies are usually seen as indicative of communication problems caused by production or self-monitoring issues \cite{levelt1983monitoring}. Individuals with AD are likely to deal with troubles in language and cognitive skills. Patients with AD speak more slowly and with longer breaks and invest extra time seeking the right word, which in effect contributes to disfluency \cite{lopez2013selection, nasreen2021alzheimer}.

We automatically annotate self-repairs and edit terms using \cite{rohanian-hough-2020-framing}'s multi-task learning model in a left-to-right, word-by-word manner to predict disfluency tags. Here each word is either tagged as one of $\{$\textit{repair onset}, \textit{edit term}, \textit{fluent word}$\}$ by the disfluency detector- we concatenate the disfluency tags with the word vectors to create the input for the text-based LSTM classifier described below.

\subsection{Unfilled Pauses}

Durations of pauses were calculated from the word timings provided by the ASR hypotheses, using the latency between the end of the previous word to the beginning of the patient's current word as the pause length, with the value for the first word being 0. We further categorized pauses into either \textit{short pause} (SP) and \textit{long pause} (LP). An SP is a silence that occurs inside a single speaker turn, which in the range $[0.5,1.5)$ seconds; an LP is a longer pause within a single speaker turn defined as a speech pause of 1.5 seconds or greater. Pauses in the interviewer's speech were excluded.

\subsection{Language Model Probabilities}

People with speech disorders or cognitive impairment express themselves in different ways when compared to control groups \cite{gabani2009corpus}. Language model probabilities, which can be interpreted to estimate the predictability of a sequence of words, can be used to assess a participant's language structure, including vocabulary and syntactic constructions. The present work uses a Multi-task Learning (MTL) LSTM language model \cite{rohanian-hough-2020-framing} based on the Switchboard corpus \cite{godfrey1992switchboard}, a sizable multispeaker corpus of conversational speech and text. The language model uses standard Switchboard training data for disfluency detection (all conversation numbers starting sw2*,sw3 * in the Penn Treebank III release: 100k utterances, 650k words) and is trained in combination with other tasks, including disfluency detection as described in \cite{rohanian-hough-2020-framing}. This corpus can be viewed as an approximation of control, non-AD disorder spoken dialogue. The model is then tested on the ASR transcript of each session, and the probability of each word is calculated. Finally, we concatenate the probability of the current word given the history $p(w_t|w_0...w_{t-1})$ with the word vectors to create the lexical input for our model.

\section{Proposed Approach}

We experiment with  different deep-learning architectures for predicting AD in both classification and regression and for cognitive decline prediction: 

\begin{itemize}
    \item[1] unimodal LSTM models utilising using either acoustic or lexical features.
    \item[2] multimodal LSTM model using lexical and acoustic information, including disfluency and pause tagging.
    \item[3] unimodal BERT based classifier using lexical features.
    \item[4] multimodal BERT model with gating using lexical and acoustic information. 
\end{itemize}

\subsection{Sequence modeling}

Our approach is to model the speech of individuals as a sequence to predict whether they have AD or not, and if so, to what degree, using either LSTMs or BERT models.

\textbf{LSTM} The potential of neural networks lies in the power to derive representations of features by non-linear input data transformations, providing greater capacity than traditional models. As we were interested in modelling the temporal nature of speech recordings and transcripts, we used a bi-directional LSTM. For each of the audio and text modalities, we trained a separate unimodal LSTM model, using different sets of features, then used late fusion to combine their probabilities.

\textbf{BERT} Pre-trained BERT models are fine-turned for the AD classification task. Each of the training instances is considered a data point. The input to the model consists of a sequence of words from the transcript for every speaker. Following \cite{yuan2020disfluencies} we used Bert-for-Sequence-Classification2 for fine-tuning. The standard default tokenizer was used, and two special tokens, [CLS] and [SEP], were added to the beginning and the end of each input. Specifically for regression, the last layer is the shape (hidden size, 1), and we use MSE loss instead of cross-entropy.

\subsection{Multimodal Model with Gating}

Since learned representation for the text can be undermined by corresponding audio representation and ASR results can be unreliable, we need to minimise the effects of noise and overlaps during multimodal fusion. For audio and textual input for the BiLSTM models, we use two branches of the LSTM, one for each of the modalities, with their outputs combined into final \textit{feed-forward highway layers} \cite{srivastava2015highway}, with gating units that learn by weighing text and audio inputs at each time step to regulate information flow through the network. 

The concatenated output is passed through N highway layers (where the best value \textit{N} was determined from optimizing on held-out data). We pad the size of the training examples in the text set (which was the smaller set) to meet the audio set by mapping together instances that occurred in the same session, as the audio and text inputs for each branch of the LSTM had different timesteps and strides.

For the BERT-based multimodal models with gating, the output from the BERT-based textual classifier is combined with the acoustic data into the final feed-forward highway layers.

\section{Experiments}

\subsection{Implementation and Metrics}

We set up our model to learn the most helpful information from modalities for predicting AD. All experiments are carried out without being conditioned on the identity of the speaker.

For the LSTM models, the sizes of layers and the learning rates are calculated by grid search on validation test.
For the input data, we explored different timesteps and strides. After exploring different hyper-parameters, the model using audio data has a timestep of 20 and stride 1 with four bi-directional LSTM layers with 256 hidden nodes. The model using text input has an input with a timestep of 10 and stride of 2 and has 2 LSTM layers with 16 hidden nodes. We use a block of 3 stacked highway layers. The LSTM models were trained using ADAM \cite{kingma2014adam} with a learning rate of 0.0001. We used Binary Cross-Entropy to model binary outcomes for the loss function and Mean Square Error (MSE) to model regression outcomes.

For the BERT models, following \cite{yuan2020disfluencies} we use the “bert-large-uncased" model, with the hyperparameters: learning rate = 2e-5, batch size = 4, epochs = 8, max input length of 256.

For binary classification of AD and non-AD, we report binary accuracy scores. For the MMSE prediction task, we report the Root Mean Square Error (RMSE) for the prediction error score. For the cognitive decline task, we report the mean of F1 classification scores.

The code used in the experiments is publicly available in an online repository.\footnote{https://github.com/mortezaro/ad-recognition-from-speech}

\subsection{Baseline Models}

We compare the performance of our models to the ADReSSo Challenge baselines \cite{bib:LuzEtAl21ADReSSo} with an ensemble of audio and linguistic features provided with the dataset. The best baselines we include here include decision trees (DT), linear discriminant analysis (LDA), support vector machines (SVM), support vector regression (SVR), and Gaussian process regression (GP).

\begin{table}[t]
\caption{Result of the AD classification and regression experiments with our models against baseline models on test set}
\label{tab:results-test}
\resizebox{\columnwidth}{!}{%
\begin{tabular}{llll}
\hline
\multicolumn{1}{|l|}{Models} & \multicolumn{1}{l|}{Features} & \multicolumn{1}{l|}{Accuracy} & \multicolumn{1}{l|}{RMSE} \\ \hline
\multicolumn{4}{l}{Baseline (\cite{bib:LuzEtAl21ADReSSo})}      
\\
LDA                          & Linguistic                      &   0.76                       & -                         \\
DT                           & Linguistic                      &       0.75                   &            6.24          \\
SVM                          & Acoustic+Linguistic                       & 0.79  &-                   \\
SVR                          & Acoustic+Linguistic                       &  - & 5.29 

\\

GP                          & Linguistic                       & -                             &          5.95            \\
\hline
\multicolumn{4}{l}{Our Models}                                     \\                               LSTM                     &     Words                  &   0.76      & -                                      \\
    LSTM                    &      Words+Words Probabilities                 &   0.77 & 4.75                                             \\
    LSTM                     &     Words+Disf+Pause              &         0.81                 &     4.43                  \\
     BERT        &   Words          &   0.80                       &    4.49   
             \\
     BERT w/ Gating      &   Words+Acoustic         &   -                       &    4.38     
     \\
     LSTM w/ Gating        &  Words+Acoustic+Disf+Pse+WP     &         \textbf{0.84}        &       \textbf{4.26}       
\end{tabular}}
\end{table}

\section{Results}

\textbf{AD classification and regression tasks} In Table~\ref{tab:results-test}, we present our proposed models' performance against that of the baselines models on AD classification and regression tasks on the provided test set and in Table~\ref{tab:results-cv} in a cross-validation setting. For AD detection, our proposed LSTM model with gating and additional features (disfluency, unfilled pause, and language model probabilities) achieves an accuracy of \textbf{0.84} and RMSE of \textbf{4.26}, outperforming all the baselines. Overall, the results support our hypothesis that a model with a gating structure can more effectively reduce individual modalities' errors and noise, including that from errorful ASR results. Furthermore, our proposed LSTM model with gating and additional features (disfluency, unfilled pauses, and language model probabilities) outperforms the BERT fine-tuned models in unimodal and multimodal situations (ACC 0.84 vs.\ 0.80; RMSE 4.26 vs.\ 4.49 and 4.38). It should also be noted that the BERT model is very large in comparison to the LSTM models. BERT has approximately 21 times the number of parameters as our second largest model (105 million vs.\ 4.9 million). Therefore, compared to the BERT model, our LSTM models need fewer resources for development.

\textbf{Effect of disfluency and unfilled pause features} We found that disfluencies and unfilled pauses help as features in AD detection. Adding disfluency and pause features to the lexical features lead to improvement on the test set (ACC 0.81 vs.\ 0.76) and in CV (ACC 0.78 vs.\ 0.74; RMSE 5.02 vs.\ 5.31). Our LSTM model with disfluencies and unfilled pauses outperforms the BERT model in both class-action and regression tasks on the test set (ACC 0.81 vs.\ 0.80; RMSE 4.43 vs.\ 4.49).

\textbf{Effect of language model probabilities} Language model probabilities (as an indicator of grammatical integrity) are useful as features in the diagnosis of AD. Adding language model probabilities to the lexical features improves the test set (ACC 0.77 vs.\ 0.76) and in CV (ACC 0.78 vs.\ 0.74; RMSE 4.78 vs. 5.31).

\begin{table}[t]
\caption{Result of the AD classification and regression experiments with our models in cross validation}
\label{tab:results-cv}
\resizebox{\columnwidth}{!}{%
\begin{tabular}{llll}
\hline
\multicolumn{1}{|l|}{Models} & \multicolumn{1}{l|}{Features} & \multicolumn{1}{l|}{Accuracy} & \multicolumn{1}{l|}{RMSE} \\ \hline
LSTM                         & Acoustic                      & 0.68                          & 6.03                      \\

LSTM                         & Words                        & 0.74                          &     5.31                  \\
LSTM                         & Words+Words Probabilities                       & 0.78                          &     4.78                  \\
LSTM                         & Words+Disfluency+Pause                  &  0.78                         &        5.02                 \\
     BERT        &   Words          &    0.80                      &       4.94
             \\
     BERT        &   Words+Acoustic         &                   0.78      &   \textbf{4.72}        
\\
LSTM w/ Gating             &      Words+Acoustic       &            0.79              &         4.88              \\
LSTM w/ Gating             & Words+Acoustic+Disf+Pse+WP    &     \textbf{0.81}            &     4.75        
\end{tabular}}
\end{table}

\textbf{Effect of multimodality} On both the test set and in CV, the multimodal LSTM with gating model outperforms the single modality AD detection models in classification and regression tasks. In CV, integrating textual and audio modalities with gating improves performance over single modality models (ACC 0.79 vs.\ 0.74; RMSE 4.88 vs.\ 5.31). Even though each LSTM branch has different steps and timestep inputs in multimodal models, adding audio features improves performance. The multimodal model with BERT outperforms the single modality BERT in the regression task on both the test set and in CV (RMSE 4.72 and 4.38 vs. \ 4.94 and 4.49). However, integrating BERT and audio model with gating decreases performance over BERT for classification in CV (ACC 0.78 vs.\ 0.80). Text features are more informative than audio features as using text modality only predicts AD better than using unimodal audio modality sequentially in CV (ACC 0.74 vs.\ 0.68; RMSE 5.31 vs.\ 6.03).

\begin{table}[!ht]
\caption{Result of Task3: cognitive decline progression results (mean of
F1Score) for leave-one-subject-out CV and Test set}
\label{tab:results-test3}
\resizebox{\columnwidth}{!}{%
\begin{tabular}{llll}
\hline
\multicolumn{1}{|l|}{Models} & \multicolumn{1}{l|}{Features} & \multicolumn{1}{l|}{CV} & \multicolumn{1}{l|}{Test} \\ \hline
\multicolumn{3}{l}{Baseline (\cite{bib:LuzEtAl21ADReSSo})}                                                                                             \\
LDA                          & Linguistic                      &   0.55                       & 0.54                         \\
DT                           & Linguistic                      &       \textbf{0.76}                   &            \textbf{0.67}         \\
SVM                          & Linguistic                       & 0.45  & 0.40                   \\
\hline
\multicolumn{4}{l}{Our Models}                                     \\                               LSTM                     &     Words                  &   0.59      &  0.55                                                                 \\
    LSTM                     &     Words+Disfluency+Pause              &         0.55                 &     0.50                  \\
     BERT        &   Words          &   0.63                       &    0.54   
     \\
     LSTM w/ Gating        &  Words+Acoustic+Disf+Pse+WP     &         0.66         &       0.62               
\end{tabular}}
\end{table}

\textbf{Cognitive decline (disease progression) inference task} In Table~\ref{tab:results-test3}, we present our results for disease progression task. As can be seen, our models do not reach the best baseline of the Decision-Tree based classifier. However, as with AD classification, the multimodal LSTM with Gating model outperforms all other competitors and is close to the DT classifier in performance on the test data (ACC \textbf{0.62} vs. 0.67). Overall, this task seems to have a considerably greater variation in performance across baseline classifiers and feature sets than the other two tasks. The lower performance of the LSTM model using words with disfluency and pause information model compared to using words alone (ACC 0.55 vs.\ 0.59) suggests these extra features are not as useful compared to the lexical information alone. This suggests the ASR quality is more critical, and the comparison of the IBM Watson system used here against the results obtained by the Google Cloud-based Speech Recogniser used by \cite{bib:LuzEtAl21ADReSSo} would be a future step to take.

\section{Conclusions}
We have presented two multimodal fusion-based deep learning models which consume ASR transcribed speech and acoustic data simultaneously to classify whether a speaker in a structured diagnostic task has Alzheimer's Disease and to what degree. Our best model, a BiLSTM with highway layers using words, word probabilities, disfluency features, pause information, and a variety of acoustic features, achieves an accuracy of 84\%. While predicting cognitive decline is more challenging, our models show improvements using the multimodal approach and word probabilities, disfluency, and pause information over word-only models. In addition, we show there are considerable gains for AD classification using multimodal fusion and gating, which can effectively deal with noisy inputs from acoustic features and ASR hypotheses.

\section{Acknowledgments}

Purver is partially supported by the EPSRC under grant EP/S033564/1, and by the European Union's Horizon 2020 programme under grant agreements 769661 (SAAM, Supporting Active Ageing through Multimodal coaching) and 825153 (EMBEDDIA, Cross-Lingual Embeddings for Less-Represented Languages in European News Media). The results of this publication reflect only the authors' views and the Commission is not responsible for any use that may be made of the information it contains.

\bibliographystyle{IEEEtran}

\bibliography{mybib}

\begin{thebibliography}{10}
\providecommand{\url}[1]{#1}
\csname url@samestyle\endcsname
\providecommand{\newblock}{\relax}
\providecommand{\bibinfo}[2]{#2}
\providecommand{\BIBentrySTDinterwordspacing}{\spaceskip=0pt\relax}
\providecommand{\BIBentryALTinterwordstretchfactor}{4}
\providecommand{\BIBentryALTinterwordspacing}{\spaceskip=\fontdimen2\font plus
\BIBentryALTinterwordstretchfactor\fontdimen3\font minus
  \fontdimen4\font\relax}
\providecommand{\BIBforeignlanguage}[2]{{%
\expandafter\ifx\csname l@#1\endcsname\relax
\typeout{** WARNING: IEEEtran.bst: No hyphenation pattern has been}%
\typeout{** loaded for the language `#1'. Using the pattern for}%
\typeout{** the default language instead.}%
\else
\language=\csname l@#1\endcsname
\fi
#2}}
\providecommand{\BIBdecl}{\relax}
\BIBdecl

\bibitem{forbes2005detecting}
K.~E. Forbes-McKay and A.~Venneri, ``Detecting subtle spontaneous language
  decline in early {A}lzheimer's disease with a picture description task,''
  \emph{Neurological Sciences}, vol.~26, no.~4, pp. 243--254, 2005.

\bibitem{bayles1982potential}
K.~A. Bayles and D.~R. Boone, ``The potential of language tasks for identifying
  senile dementia,'' \emph{Journal of Speech and Hearing Disorders}, vol.~47,
  no.~2, pp. 210--217, 1982.

\bibitem{kave2003morphology}
G.~Kav{\'e} and Y.~Levy, ``Morphology in picture descriptions provided by
  persons with {A}lzheimer's disease,'' \emph{Journal of Speech, Language, and
  Hearing Research}, 2003.

\bibitem{szatloczki2015speaking}
G.~Szatloczki, I.~Hoffmann, V.~Vincze, J.~Kalman, and M.~Pakaski, ``Speaking in
  alzheimer’s disease, is that an early sign? importance of changes in
  language abilities in alzheimer’s disease,'' \emph{Frontiers in aging
  neuroscience}, vol.~7, p. 195, 2015.

\bibitem{weiner2016speech}
J.~Weiner, C.~Herff, and T.~Schultz, ``Speech-based detection of alzheimer's
  disease in conversational german.'' in \emph{INTERSPEECH}, 2016, pp.
  1938--1942.

\bibitem{ambrosini2019automatic}
E.~Ambrosini, M.~Caielli, M.~Milis, C.~Loizou, D.~Azzolino, S.~Damanti,
  L.~Bertagnoli, M.~Cesari, S.~Moccia, M.~Cid \emph{et~al.}, ``Automatic speech
  analysis to early detect functional cognitive decline in elderly
  population,'' in \emph{2019 41st Annual International Conference of the IEEE
  Engineering in Medicine and Biology Society (EMBC)}.\hskip 1em plus 0.5em
  minus 0.4em\relax IEEE, 2019, pp. 212--216.

\bibitem{jarrold2014aided}
W.~Jarrold, B.~Peintner, D.~Wilkins, D.~Vergryi, C.~Richey, M.~L.
  Gorno-Tempini, and J.~Ogar, ``Aided diagnosis of dementia type through
  computer-based analysis of spontaneous speech,'' in \emph{Proceedings of the
  Workshop on Computational Linguistics and Clinical Psychology: From
  Linguistic Signal to Clinical Reality}, 2014, pp. 27--37.

\bibitem{abel2009connectionist}
S.~Abel, W.~Huber, and G.~S. Dell, ``Connectionist diagnosis of lexical
  disorders in aphasia,'' \emph{Aphasiology}, vol.~23, no.~11, pp. 1353--1378,
  2009.

\bibitem{gosztolya2019identifying}
G.~Gosztolya, V.~Vincze, L.~T{\'o}th, M.~P{\'a}k{\'a}ski, J.~K{\'a}lm{\'a}n,
  and I.~Hoffmann, ``Identifying mild cognitive impairment and mild
  alzheimer’s disease based on spontaneous speech using asr and linguistic
  features,'' \emph{Computer Speech \& Language}, vol.~53, pp. 181--197, 2019.

\bibitem{rohanian2019detecting}
M.~Rohanian, J.~Hough, M.~Purver \emph{et~al.}, ``Detecting depression with
  word-level multimodal fusion,'' \emph{Proc. Interspeech 2019}, pp.
  1443--1447, 2019.

\bibitem{Rohanian2020}
\BIBentryALTinterwordspacing
M.~Rohanian, J.~Hough, and M.~Purver, ``{Multi-Modal Fusion with Gating Using
  Audio, Lexical and Disfluency Features for Alzheimer’s Dementia Recognition
  from Spontaneous Speech},'' in \emph{Proc. Interspeech 2020}, 2020, pp.
  2187--2191. [Online]. Available:
  \url{http://dx.doi.org/10.21437/Interspeech.2020-2721}
\BIBentrySTDinterwordspacing

\bibitem{bib:LuzEtAl21ADReSSo}
S.~Luz, F.~Haider, S.~de~la Fuente, D.~Fromm, and B.~MacWhinney, ``Detecting
  cognitive decline using speech only: The adresso challenge,'' \emph{medRxiv},
  2021.

\bibitem{benton1968differential}
A.~L. Benton, ``Differential behavioral effects in frontal lobe disease,''
  \emph{Neuropsychologia}, vol.~6, no.~1, pp. 53--60, 1968.

\bibitem{degottex2014covarep}
G.~Degottex, J.~Kane, T.~Drugman, T.~Raitio, and S.~Scherer, ``Covarep—a
  collaborative voice analysis repository for speech technologies,'' in
  \emph{2014 ieee international conference on acoustics, speech and signal
  processing (icassp)}.\hskip 1em plus 0.5em minus 0.4em\relax IEEE, 2014, pp.
  960--964.

\bibitem{baumann2017recognising}
T.~Baumann, C.~Kennington, J.~Hough, and D.~Schlangen, ``Recognising
  conversational speech: What an incremental asr should do for a dialogue
  system and how to get there,'' in \emph{Dialogues with social robots}.\hskip
  1em plus 0.5em minus 0.4em\relax Springer, 2017, pp. 421--432.

\bibitem{pennington2014glove}
J.~Pennington, R.~Socher, and C.~D. Manning, ``Glove: Global vectors for word
  representation,'' in \emph{Proceedings of the 2014 conference on empirical
  methods in natural language processing (EMNLP)}, 2014, pp. 1532--1543.

\bibitem{levelt1983monitoring}
W.~J. Levelt, ``Monitoring and self-repair in speech,'' \emph{Cognition},
  vol.~14, no.~1, pp. 41--104, 1983.

\bibitem{lopez2013selection}
K.~L{\'o}pez-de Ipi{\~n}a, J.-B. Alonso, C.~M. Travieso, J.~Sol{\'e}-Casals,
  H.~Egiraun, M.~Faundez-Zanuy, A.~Ezeiza, N.~Barroso, M.~Ecay-Torres,
  P.~Martinez-Lage \emph{et~al.}, ``On the selection of non-invasive methods
  based on speech analysis oriented to automatic alzheimer disease diagnosis,''
  \emph{Sensors}, vol.~13, no.~5, pp. 6730--6745, 2013.

\bibitem{nasreen2021alzheimer}
S.~Nasreen, M.~Rohanian, M.~Purver, and J.~Hough, ``Alzheimer's dementia
  recognition from spontaneous speech using disfluency and interactional
  features,'' \emph{Frontiers in Computer Science}, vol.~3, p.~49, 2021.

\bibitem{rohanian-hough-2020-framing}
\BIBentryALTinterwordspacing
M.~Rohanian and J.~Hough, ``Re-framing incremental deep language models for
  dialogue processing with multi-task learning,'' in \emph{Proceedings of the
  28th International Conference on Computational Linguistics}.\hskip 1em plus
  0.5em minus 0.4em\relax Barcelona, Spain (Online): International Committee on
  Computational Linguistics, Dec. 2020, pp. 497--507. [Online]. Available:
  \url{https://www.aclweb.org/anthology/2020.coling-main.43}
\BIBentrySTDinterwordspacing

\bibitem{gabani2009corpus}
K.~Gabani, M.~Sherman, T.~Solorio, Y.~Liu, L.~Bedore, and E.~Pena, ``A
  corpus-based approach for the prediction of language impairment in
  monolingual english and spanish-english bilingual children,'' in
  \emph{Proceedings of human language technologies: the 2009 annual conference
  of the North American chapter of the Association for Computational
  Linguistics}, 2009, pp. 46--55.

\bibitem{godfrey1992switchboard}
J.~J. Godfrey, E.~C. Holliman, and J.~McDaniel, ``Switchboard: Telephone speech
  corpus for research and development,'' in \emph{Acoustics, Speech, and Signal
  Processing, IEEE International Conference on}, vol.~1.\hskip 1em plus 0.5em
  minus 0.4em\relax IEEE Computer Society, 1992, pp. 517--520.

\bibitem{yuan2020disfluencies}
J.~Yuan, Y.~Bian, X.~Cai, J.~Huang, Z.~Ye, and K.~Church, ``Disfluencies and
  fine-tuning pre-trained language models for detection of alzheimer’s
  disease,'' \emph{Proc. Interspeech 2020}, pp. 2162--2166, 2020.

\bibitem{srivastava2015highway}
R.~K. Srivastava, K.~Greff, and J.~Schmidhuber, ``Highway networks,''
  \emph{arXiv preprint arXiv:1505.00387}, 2015.

\bibitem{kingma2014adam}
D.~P. Kingma and J.~Ba, ``Adam: A method for stochastic optimization,''
  \emph{arXiv preprint arXiv:1412.6980}, 2014.

\end{thebibliography}


\end{document}